\documentclass{article}
\usepackage{spconf,amsmath,graphicx}
\usepackage{color}
\usepackage{hyperref}
\usepackage{bm}
\usepackage{graphicx}
\usepackage[normalem]{ulem}
\usepackage{amsmath}
\usepackage{bbm}
\usepackage{amssymb}
\usepackage{fullpage}
\usepackage{setspace}
\usepackage{array}
\usepackage{rotating}
\usepackage[section]{placeins}
\usepackage{url}
\usepackage[utf8]{inputenc}
\usepackage{algorithm}
\usepackage{algorithmic}
\usepackage{booktabs}
\usepackage{booktabs}  

\newcommand{\be}{\begin{equation}}
\newcommand{\ee}{\end{equation}}
\newcommand{\bea}{\begin{eqnarray}}
\newcommand{\eea}{\end{eqnarray}}

\title{Hyper-parameter optimization of Deep Convolutional Networks for object recognition}
\name{Sachin S. Talathi}
\address{Qualcomm Research Center, 5775 Morehouse Dr, San Diego CA 92121}
\begin{document}
%
\maketitle
\begin{abstract}
Recently sequential model based optimization (SMBO) has emerged as a promising hyper-parameter optimization strategy in machine learning.  In this work, we investigate SMBO to identify architecture hyper-parameters of deep convolution networks (DCNs) object recognition. We propose a simple SMBO strategy that starts from a set of random initial DCN architectures to generate new architectures, which on training perform well on a given dataset. Using the proposed SMBO strategy we are able to identify a number of DCN architectures that produce results that are comparable to state-of-the-art results on object recognition benchmarks.

\end{abstract}
\section{Introduction}
\label{sec:intro}
The primary task for a supervised machine learning algorithm is to use training dataset \{$x_{tr},y_{tr}$\} to find a function $f:x\rightarrow y$, that also generalize well across the test (or the hold out) dataset \{$x_{h},y_{h}$\}.  Very often $f$ is obtained through the optimization of a training criterion, $C$, with respect to a sets of parameters, $\theta$. The learning algorithm used to optimize $C$ usually contains its own set of free parameters $\lambda_{l}$, referred to as the learning algorithm hyper-parameters. These hyper-parameters are often estimated using grid search cross validation. In addition to the learning algorithm hyper-parameters, $\lambda_{l}$, neural network models such as deep convolution networks (DCNs) also comprise of hyper-parameters $\lambda_{a}\notin \lambda_{l}$, that define the architectural configuration of the network. Grid search techniques are prohibitively expensive to tune $\lambda_{a}$, given the fact that there are a few tens of these architectural hyper-parameters. As a result, many of the state-of-the-art DCNs are manually designed, making the task of tuning these hyper-parameters more of an art than a science \cite{Bergstra_2011}. 

In recent years, there has been a concerted effort in the machine learning community to develop better algorithms to solve the hyper-parameter optimization problem \cite{Bergstra_2011, hyperopt, spear,Brochu_2010,Bengio_2012}. Many of these works have successfully applied direct search methods for non-linear optimization such as the sequential model based  optimization (SMBO) to generate better results on various supervised machine learning tasks than were previously reported. Motivated by these works, in this paper we attempt to address the question: {\it Can SMBO be used to determine superior architectural configurations for DCNs?}  The paper is organized as follows: In the Methods Section 2, we will briefly formulate the problem of hyper-parameter optimization for DCNs. We then present the general strategy of sequential model based optimization (SMBO) \cite{Brochu_2010} and summarize our approach to SMBO for designing DCN architectures. The results of image classification training and evaluation on the benchmark CIFAR-10 dataset are then presented in the Results Section 3, which is followed by the Conclusion.

\section{Methods}
\subsection{Formulation of the problem}
Let $M_{\lambda}(x,w)$ represent the DCN model that operates on input data $x \in \mathbb{R}^{D}$ and generates an estimate $\hat{y}$ for the output data $y\in\mathbb{Z}_{2}^{C}$ . The DCN model, $M$, is parameterized by two set of parameters, the first being $w$, which are obtained through the optimization of a training criterion, $C$, using a gradient descent type learning algorithm such as the back-propagation algorithm and the second being $\lambda=\{\lambda_{a},\lambda_{l}\}$, which represent the set of the so called hyper-parameters. The hyper-parameters $\lambda_{a}$ define the DCN architecture and the hyper-parameters $\lambda_{l}$, are associated with the learning algorithm used to optimize $C$.
The objective for DCN hyper-parameter optimization is to solve the joint optimization problem as stated below:
\bea
\{w,\lambda\}&=&\underset{\lambda}{\operatorname{argmin}}\left[\Psi(\hat{y}_{h},y_{h})\right] \quad \text{where,}\nonumber \\
\hat{y}_{h}&=& M_{\lambda}(\underset{w}{\operatorname{argmin}}\left(C(x_{tr},y_{tr},\theta)\right),x_{h})
\eea

where \{$(x_{tr},y_{tr}),(x_{h},y_{h})$\}$\in (x,y)$, are the input and output training and hold-out (or the test) data set respectively, $\Psi=\sum_{\{y_{h}\}}\mathbb{I}_{y_{h}\ne\hat{y}_{h}}$. 

\subsection{Sequential model based optimization}
SMBO is a direct search method for non-linear optimization, in which one begins by selecting a meta-model of the function for which an extrema is sought. One then applies  an active learning strategy to select a query point that provides the most potential to approach the extrema. More specifically, let us assume that we have a database $D_{1:t}=\{\lambda_{1:t},e_{1:t}\}$ of $t$ DCN models, where $\lambda_{i}|_{i=1}^{t}$ represents the set of hyper-parameters and $e_{i}|_{i=1}^{t}$ is the validation error on the hold-out dataset generated by each of the $t$ DCN models. The basic idea underlying SMBO is to replace the original optimization problem of finding extrema of a given function, such as $\Psi(\lambda)$, which is time consuming and computationally expensive, with an equivalent problem of optimization of expected value of an utility function, $u(e)$ \cite{Brochu_2010}. As we will see below, optimizing over the expected value of the utility function is computationally much cheaper and faster than solving the original problem. Under SMBO, one usually begins by assigning a prior distribution $p(e)$ on $e$. One then uses the database $D_{1:t}$ to obtain an estimate for the likelihood function $p(\lambda_{1:t}|e)$. The prior and the likelihood function are used to obtain an estimate for the posterior distribution $p(e|\lambda_{1:t})\propto p(\lambda_{1:t}|e)p(e)$. The objective then is to choose $\lambda_{t+1}$, which maximizes u(e) under $p(e|\lambda_{1:t})$, i.e., $\lambda_{t+1}=\underset{\lambda}{\operatorname{argmax}}\left[E(u(e))\right]$, where $E(u(e))$ is given as:
 \bea
 E(u(e))&=&\int u(e) p(e|\lambda_{1:t})  de \nonumber \\
&=&\int u(e) \frac{p(\lambda_{1:t}|e)p(e)}{p(\lambda_{1:t})} de
 \eea
 
 A common choice for the utility function $u(e)$, is the Expected Improvement function \cite{Brochu_2010}, $u(e)=\max\left((e^{*}-e),0\right)$, in which case, Eq. 4 becomes
 \bea
 E(u(e))&=&\int_{0}^{e^{*}} (e^{*}-e) p(e|\lambda_{1:t}) de
 \eea
Then under SMBO we have, 
\bea \lambda_{t+1}=\underset{\lambda}{\operatorname{argmax}}\left[\frac{e^{*}}{p(\lambda_{1:t})}\int_{0}^{e^{*}} p(\lambda_{1:t}|e)p(e) de\right]\eea
If $p(e<e^{*})=\gamma$ (a constant), i.e., choose $e^{*}$ to be some quantile of observed $e$ values, and we define two density functions; $l(\lambda)=p(\lambda_{1:t}|e)$ when $e\le e^{*}$ and $g(\lambda)=p(\lambda_{1:t}|e)$ when $e> e^{*}$ as proposed in \cite{Bergstra_2011}, then, \bea \lambda_{t+1}=\underset{\lambda}{\operatorname{argmax}}\left [\frac{e^{*}\gamma l(\lambda)}{\gamma l(\lambda)+(1-\gamma)g(\lambda)}\right]\eea

Bergstra et.al., $\cite{Bergstra_2011}$, proposed an adaptive Parson estimator algorithm to evaluate Eq. 5 so as to maximize the ratio $l(\lambda)/g(\lambda)$. In this work, we propose a simple strategy to evaluate Eq. 5 as follows: let $l(\lambda)+g(\lambda)=U(\lambda)=k$ (a constant). In other words, $\lambda$ is drawn from a uniform distribution. Then, if $\gamma=0.5$, we have from Eq. 5 \bea \lambda_{t+1}=\frac{2e^{*}}{k}\underset{\lambda}{\operatorname{argmax}}\left [ l(\lambda)\right]\eea
Our proposed simplification in Eq. 6, does not require an estimate for both $l(\lambda)$ and $g(\lambda)$. We can simply pick $\lambda\in U(\lambda)$ and evaluate according to empirical distribution of $l(\lambda)$ generated from $D_{1:t}$ to evaluate the next potential $\lambda_{t+1}$. Since the proposed algorithm chooses $\lambda\in U(\lambda)$ at every step, a much larger space of potential $\lambda$'s are explored, which may in turn slow down the convergence rate to an optimal $\lambda^{*}$. In order to counter this,we adopt a hybrid strategy by combining Eq. 5 and Eq. 6. With probability $p$, at every iteration, we choose the potential $\lambda_{t+1}$ according to Eq. 5 and with probability $1-p$, we choose the potential $\lambda_{t+1}$ according to Eq. 6. In Table 1, we summarize the steps involved in our proposed algorithm.

\begin{table}
\begin{minipage}{8 cm }
{\bf HyperOpt} ($D_{1:t},p,N$):\\ 
 {\bf  While } $t\le N$ {\bf do}:\\
1. Estimate $e^{*}$ s.t. $p(e_{1:t}<e^{*})=0.5$\\
2. Use $e^{*}$ and $\lambda_{1:t}$ to estimate $l(\lambda)$ and $g(\lambda)$\\
3. Evaluate $\lambda_{t+1}$ according to Eq. 5 and Eq. 6\\
4. Train the new DCN model with $\lambda_{t+1}$ to estimate $e_{t+1}$ 
5. $t\leftarrow  t+1$
\caption{SMBO algorithm for estimating architecture hyper-parameters for DCN.\label{Table2}}
\end{minipage}
\end{table}

\section{Results}
We evaluate our proposed SMBO algorithm on the CIFAR-10 benchmark, which consists of 60,000 32x32 color images. The collection is split into 50,000 training and 10,000 testing images. All the DCNs generated by our proposed algorithm were trained using cuda-convnet2\footnote{https://code.google.com/p/cuda-convnet2/}. We used a 3 step cooling procedure; starting with learning rate $l=0.01$, the momentum $m=0.9$, the weight decay parameter $w_{c}=0.0005$ for the first 120 epochs followed by another 20 epochs by reducing learning rate by a factor of 10 (keeping other parameters the same) and then training for 10 more epochs by further reducing learning rate by a factor of 10.  

Since the primary focus for us in this work is to determine whether SMBO can be used to identify suitable DCN architectures, we fixed the DCN hyper-parameters associated with the back-propagation learning algorithm as described above. The set of DCN architecture hyper-parameters that we consider for optimization are listed in Table 2.
\begin{table}[h]
\begin{tabular}{|l|l|lll}
\cline{1-2}
Conv. Layer  & \begin{tabular}[c]{@{}l@{}}1. No. of Conv. layers\\ 2. No. of filters per layer\\ 3. Filter size; 4. Filter stride\end{tabular} &  &  &  \\ \cline{1-2}
Norm. Layer  & 1. Size                                                                                                                         &  &  &  \\ \cline{1-2}
Pool Layer   & \begin{tabular}[c]{@{}l@{}}1. Size; 2. Stride\\ 3. type of pooling: max/avg.\end{tabular}                                       &  &  &  \\ \cline{1-2}
Hidden Layer & \begin{tabular}[c]{@{}l@{}}1. No. of hidden layers\\ 2. No. of nodes per hidden layer\\ 3. dropout value\end{tabular}           &  &  &  \\ \cline{1-2}
\end{tabular}
\caption{DCN architecture hyper-parameters. In addition to the parameters listed above; we also consider two additional boolean hyper-parameters to represent the presence or absence of the Norm layer or the pool layer and a third boolean hyper-parameter indicating the presence or absence of dropout. For normalization layer; we only consider local response normalization across filter maps \cite{Alex}, with a scaling factor of 0.75.}
\end{table}

It has been reported in the literature \cite{Simonyan_2014} that very deep networks are difficult to train primarily suffering from vanishing gradient problem at larger depths. In order to alleviate this problem, for our implementation of SMBO for DCN architectural hyper-parameters, all the DCNs  are generated to comprise a local logistic regression (LR) cost function layer at the output of one or more of the convolution block.  

For the results presented here, we consider $t=32$ as the size of our initial database based on our analysis of random search hyper-parameter optimization \cite{Bengio_2012} and we set $p=0.9$.

\begin{table*}[t]
\resizebox{\textwidth}{!}{
\begin{tabular}{lllllllllllllll}
\toprule
& & \multicolumn{8}{c}{\bf Convolutions} & \multicolumn{3}{c}{\bf Fully-connected} &  \multicolumn{2}{c}{\bf Test error}\\ 
\cmidrule(lr){3-10} \cmidrule(lr){11-13} \cmidrule(lr){14-15}
Architecture & \# Trainable Parameters & Parameters 		& Layer 1 	& Layer 2 & Layer 3 & Layer 4 & Layer 5 & Layer 6 & Layer 7 & Layer 8 & Layer 1 & Layer 2 & Softmax	& 550 epochs \\
\midrule
{\bf DCN 1.} &$\approx$30.9 M		& Filter / size		& 64x3x3	& 256x3x3 & 256x3x3 & 256x3x3 & 256x3x3 & 256x3x3 & 256x3x3 & 256x11x11 & 3,314 	& 4,951	  & 10	& 7.81\% \\
			&& Stride			& 1			& 1		  & 2		& 2		  & 1		& 2		  & 2		& 10		  &  		&  & &	  	  	\\
			&& Padding			& 0			& 0		  & 1		& 1		  & 0		& 1		  & 1		& 9		  &  		& 	  &  &	  	\\
			&& Pooling	(Size, Stride)	& 			& 	(2,2)	  & 		& 		  & 		& 		  & 		& 	  &  		& 	 & & 	  	\\
			&& Normalization		& 			& 		  & 		& 		  & \checkmark		& 		  & \checkmark		& \checkmark	   		& 	  	  &  &  &	\\
			&& Dropout			& 			& 		  & 		& 		  & 		& 		  & 		& 		  & \checkmark & \checkmark  &  & \\
\midrule
{\bf DCN 2.}& $\approx$4.0 M& Filter / size		& 128x3x3	& 128x3x3 & 128x3x3 & 256x3x3 & 256x3x3 & 256x7x7 & NA 		& NA	  &  	NA	& 	NA  	  & 10  &  8.17\% \\
			&& Stride			& 1			& 2		  & 1		& 1		  & 1		& 2		  & 		& 		  &  		& 	  	  &  &    \\
			&& Padding			& 0			& 1		  & 0		& 0		  & 0		& 1		  & 		& 		  &  		& 	  	  &  &   \\
			&& Pooling (Size, Stride)		& 		&        & 		& 		  & 		& 	  & 		&		  &  		& 	  	  &     &\\
			&& Normalization		& 		& \checkmark       & 		& 		  & \checkmark		& \checkmark	  & 		&		  &  		& 	  	  &  &    \\		
			&& Dropout			& 			& 		  & 		& 		  & 		& 		  & 		& 	  	  &  		& 		  &  	\\	
\midrule
{\bf DCN 3.}&$\approx$3.4 M& Filter / size		& 256x3x3	& 128x3x3 & 256x3x3 & 256x3x3 & 256x3x3 &  128x7x7 & NA 		& NA	  &  	NA	& 	NA  	  & 10	 & 8.63\% \\
			&& Stride			& 1			& 2		  & 1		& 1		  & 2		& 5		  & 		& 		  &  		& 	  	  &  &	\\
			&& Padding			& 0			& 1		  & 0		& 0		  & 1		& 2		  & 		& 	      &  		& 	  	  & 	&\\
			&& Pooling (Size,Stride)		& 		& 		  & 		& 		  & 		& 	  & 		&		  &  		& 	  	  & 	&\\
			&& Normalization		&\checkmark 		& 		  & 		& 		  & 		& 	  & 		&		  &  		& 	  	  &  &	\\
			&& Dropout			& 			& 		  & 		& 		  & 		& 		  & 		& 		  &         &        & & 	\\
\bottomrule
\end{tabular}
}
 \caption{Architecture for the top 3 DCNs generated by our proposed SMBO algorithm.\label{Table2}}
\end{table*}

\begin{figure}[]
  \centering
    \includegraphics[scale=0.4]{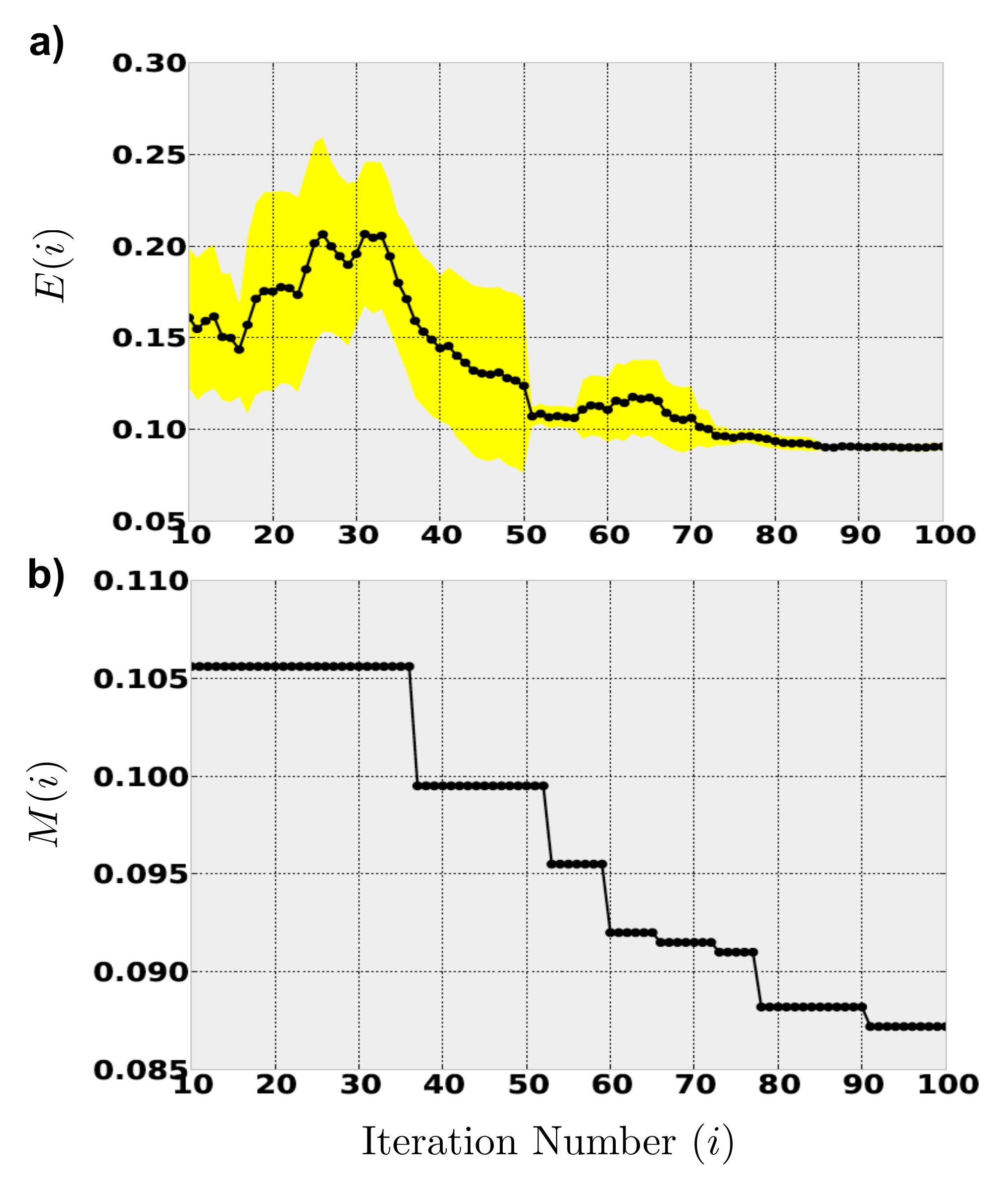}
  \caption{ (a) The mean test error and standard deviation (in yellow) as function of the SMBO iteration number for multi-view mode (b) The minimum multi-view mode test error as function of SMBO iteration number  \label{Fig1}}
\end{figure}

In Figure \ref{Fig1}a, we plot the mean (std. error, shown in yellow) test error (evaluated in multi-view test mode, \cite{Alex})  $E(i)=\left<e_{i-10:i}\right>$ and in Figure \ref{Fig1}b, we plot the minimum test error $M(i)=\min(e_{0:i})$ as function of the iteration number i, respectively. We see that the average test error gradually decrease towards an optimal solution, the best minimum found also decreases with increasing iterations. Furthermore, our proposed SMBO procedure generated a large number of ``good" DCN architectures that produce test-error of $<11$ \% even with only 150 training epochs (not-reported). In comparison, the best hand-tuned DCN architecture, produced by \cite{Alex} exhibits 11\% test error in multi-view mode and requires a longer training time on the order of 500 epochs. 

Since the state-of-the-art performance numbers for the CIFAR-10 benchmark dataset are usually reported in multi-view mode (with data-augmentation \cite{Alex}), we report  multi-view test error of $7.81$\% for the best DCN generated by our proposed hyper-parameter optimization strategy, which compares favorably to the current state-of-the-art result on CIFAR-10 of $7.97$\% \cite{DSN}.   In Table \ref{Table2}, we summarize the top 3 DCN architectures found by our proposed SMBO procedure that produced multi-view test error $<9$\% on the CIFAR-10 benchmark. In \ref{Table3}, we summarize the number of parameters in each of these 3 DCN models.

At the time of writing of this manuscript for camera ready version, we came across a recent paper \cite{Dos_2015}, that reported multi-view test error of $7.25$\% on CIFAR-10 benchmark, using a hand designed DCN network, that is comprised of only convolution layers and has 1.3 M parameters. Yet another paper \cite{He_2015}, reported the utility of using parametric-relu neuron as opposed to the relu neurons. While none of the optimized DCN networks that we report in Table \ref{Table2} generate better performance numbers than the latest state-of-the-art numbers reported in \cite{Dos_2015}, we wanted to determine whether the use of parametric-relu neuron can boost the performance of the optimized DCN networks that we have identified through the hyper-parameter optimization approach. Accordingly, we retrained the smallest of the three DCN networks from Table \ref{Table2} using a version of parametric-rectified non-linear neurons of type $y=ax(x\le0)+\sqrt{x}(x>0)$, where $a$ is a learnable parameter, fixed per neuron layer in the DCN network.  We were able to obtain multi-view test error score of 6.9 \%, which to the best of our knowledge, represents the state-of-the-art score for CIFAR-10 benchmark. In Table \ref{Table4}, we summarize all the known best in class numbers for CIFAR-10 benchmark.

\begin{table}[h]
\resizebox{\columnwidth}{!}{
\begin{tabular}{llll}
\hline
\multicolumn{4}{l}{{\bf CIFAR-10 Classification error} (with data augmentation)}                                                                           \\ \hline
\multicolumn{1}{|l|}{Method} & \multicolumn{1}{l|}{Activation Function Type} & \multicolumn{1}{l|}{Error \%} & \multicolumn{1}{l|}{Trainable Params} \\ \hline
Maxout   \cite{Goodfellow_2013}                    & maxout                                        & 9.38                          & \textgreater 6 M                      \\
Dropconnect \cite{Li_2013}                 & relu                        & 9.32                          & -                                     \\
dasNet \cite{Stollenga_2013}                      & maxout                                        & 9.22                          & \textgreater 6 M                      \\
Network in Network  \cite{Lin_2014}         & relu                         & 8.81                          & $\approx$1 M                                   \\
Deeply Supervised \cite{Lee_2014}           & relu                         & 7.97                          & $\approx$1 M                                   \\
All-CNN   \cite{Dos_2015}                   & relu                         & 7.25                          & $\approx$1.3 M                                 \\
DCN-3  (Ours)                      & p-renu           & 6.9                           & $\approx$3.4 M                                 \\ \hline
\end{tabular}
}
\caption{Comparison of state-of-the-art results for CIFAR-10 benchmark; relu: rectified linear unit; p-renu: parametric-rectified non-linear unit. \label{Table4}}
\end{table}

\section{Conclusion} 
In this paper, we have proposed a simple SMBO algorithm and a recipe for hyper-parameter optimization of DCN architectures. We have demonstrated that SMBO can be used to generate a large number of  ``good" DCN architectures, which may then form a backbone for further investigations. Our results suggest that indeed SMBO can be used to identify superior DCNs. In summary, our work in this paper in addition to those from earlier works \cite{Bergstra_2011, spear} broaden the scope of the models that can be realistically investigated, without the need for the researchers to be restricted to manual evaluation of a few architectural parameters at any given time. 

\bibliographystyle{IEEEbib}
\bibliography{references}

\end{document}